\documentclass{article}
\usepackage{arxiv}

\usepackage[utf8]{inputenc} 
\usepackage[T1]{fontenc}    
\usepackage{hyperref}       
\usepackage{url}            
\usepackage{booktabs}       
\usepackage{amsfonts}       
\usepackage{nicefrac}       
\usepackage{microtype}      
\usepackage{lipsum}
\usepackage{mathtools}
\usepackage{graphicx}
\usepackage{mathrsfs}
\usepackage{amsmath}
\usepackage{amssymb}
\usepackage{natbib}
\usepackage{soul,color}

\DeclareMathOperator{\EX}{\mathbb{E}}

\newcommand{\comment}[1]{}

\newtheorem{remark}{Remark}
\newtheorem{lemma}{Lemma}

\title{Interpretable Deep Gaussian Processes with Moments}

\author{Chi-Ken Lu
        \And Scott Cheng-Hsin Yang
        \And Xiaoran Hao
        \And Patrick Shafto}

\date{
    Mathematics and Computer Science, Rutgers University Newark\\[2ex]
    \today
}

\begin{document}

\maketitle

%

%





\begin{abstract}

Deep Gaussian Processes (DGPs) combine the expressiveness of Deep Neural Networks (DNNs) with quantified uncertainty of Gaussian Processes (GPs). Expressive power and intractable inference both result from the non-Gaussian distribution over composition functions. We propose interpretable DGP based on approximating DGP as a GP by calculating the exact moments, which additionally identify the heavy-tailed nature of some DGP distributions. Consequently, our approach admits interpretation as both NNs with specified activation functions and as a variational approximation to DGP. We identify the expressivity parameter of DGP and find non-local and non-stationary correlation from DGP composition. We provide general recipes for deriving the effective kernels for DGP of two, three, or infinitely many layers, composed of homogeneous or heterogeneous kernels. Results illustrate the expressiveness of our effective kernels through samples from the prior and inference on simulated and real data and demonstrate advantages of interpretability by analysis of analytic forms, and draw relations and equivalences across kernels.




\end{abstract}

\section{INTRODUCTION}

The success of deep learning models is generally perceived as stemming from greater expressivity which results in powerful generalization~\citep{goodfellow2016deep}. However, deep learning models are viewed as black boxes because their complexity arises from the enormous number of parameters and possible choices for different structures and activation units. Understanding these models remains an open and challenging problem~\citep{zhang2016understanding,belkin2018understand,mei2018mean,advani2013statistical,liao2018dynamics}. \cite{williams1997computing} demonstrated that the characteristics of single-layer neural networks (NNs) can be understood from its effective kernel showing non-stationary and non-local correlation. It is thus appealing to create more interpretable methods through the correspondence between deep learning and kernel-based methods which have the advantage of an explicit mathematical formalization~\citep{scholkopf1998nonlinear,rahimi2008random,cho2009kernel,mairal2014convolutional,wilson2014fast,wilson2016deep,van2017convolutional,sun2018differentiable}. 

Quantifying uncertainty of inferences is also a critical issue for deep learning models~\citep{wang2013fast,gal2016dropout}. Gaussian processes (GPs)~\citep{rasmussen2006gaussian}, the infinitely wide limit of single-layer neural network with random weight parameters~\citep{neal2012bayesian}, are attractive alternative models capable of quantifying uncertainty through Bayesian inference. Moreover, GPs can be composed into DGPs~\citep{damianou2013deep}, with inference via variational approximations~\citep{titsias2010bayesian,bui2016deep,cutajar2017random,salimbeni2019deep,salimbeni2017doubly} and sampling-based method~\citep{havasi2018inference}, to achieve expressiveness that is comparable to DNNs. However, as for DNNs, with this expressivity comes difficulty in interpretability. It is desirable to leverage interpretability of explicit mathematical formalizations associated with kernel-based methods, while preserving expressivity and uncertainty quantification.

A brief review on GP and intractability of DGP is given in Sec.~\ref{sec:review}. The multi-modal posterior distribution is shown to result from latent function symmetry in the nested probabilistic model. We introduce calculation of the second and fourth moments in Sec.~\ref{sec:second_moment} and \ref{sec:fourth_moment}, respectively. With the second moment, we approximate DGP as GP in Sec.~\ref{sec:approximation} where the validity and the heavy-tailed nature of DGP are discussed by examining the fourth moments. This can be regarded as a new variational approximation to Bayesian inference in DGPs that effectively represents any depth as single layer. In Sec.~\ref{sec:interpretation}, with these analytic effective covariance functions, we interpret DGP as generation of non-local, non-stationary, and multi-scale correlation through depth of homogeneous or heterogeneous function composition, which leads to the expressiveness of DGP. Sec.~\ref{sec:optimization} demonstrates the expressivness of DGP with optimized marginal likelihood for three-layer DGP against the expressivity parameter defined in DNN model.



\subsection{Related Works}

DGP~\citep{damianou2013deep} and the earlier Warped GP~\citep{snelson2004warped,lazaro2012bayesian} were proposed to generalize the idea of GP to serve as a prior distribution over composition function in Bayesian learning framework. The need for approximate inference presents challenges, however. Using variational approach~\citep{titsias2009variational,titsias2010bayesian,salimbeni2017doubly} and expectation propagation~\citep{minka2001expectation,bui2016deep} DGPs have demonstrated superior expressivity over GPs. 
Our contribution is the development of Bayesian DGPs with analytic forms for compositional homogeneous and heterogeneous kernels as well as identification of heavy-tailed distribution associated with some DGP models. 
Because activation functions of DNNs correspond with kernels in DGPs, our results shall also help interpret the heterogeneous networks using different activation units. 

\cite{williams1997computing} was first to derive analytic kernels of one-layer neural networks, using sigmoidal and Gaussian activation functions. Later, \cite{cho2009kernel} extended analytic results to polynomial activation functions and further made extension to deep models. \cite{duvenaud2014avoiding} studied pathologies in DGPs, primarily focusing on squared exponential kernels and sampling functions from the prior; \cite{dunlop2018deep} made connection between the effective depth and ergodicity.
\cite{Daniely2016TowardDU} extended these results for homogeneous deep networks by proposing the dual activation for the Hermite, step, and exponential functions. \cite{Poole2016ExponentialEI} studied the DNN models and found expressivity parameter indicating chaotic phase and ordered phase from the correlation map. 
Our approach differs by focusing on DGPs for Bayesian inference, and allowing heterogeneous kernels.

\cite{lee2017deep} extended the correspondence between NNs and GPs, established by \cite{neal2012bayesian}, to DNNs and GPs, via the central limit theorem, focusing on homogeneous kernels.  

\section{GP and DGP Overview}\label{sec:review}

A Gaussian process (GP) is a prior distribution $p(f)$ over continuous function $f$ of which any subset of points $\{f_i\}$ follows the joint multivariate Gaussian distribution specified by mean function $\mu({\bf x}_i)$ and covariance matrix $K_{ij}=k({\bf x}_i,{\bf x}_j)$. The properties of function, such as smoothness, are encoded in the covariance function $k({\bf x}_i,{\bf x}_j):{ R}^D\times{ R}^D\rightarrow R$. For example, functions that are sampled from the squared exponential (SE) kernel are infinitely differentiable, while the non-stationary neural network kernel~\citep{williams1997computing} may generate functions which resemble step functions. Under the Bayesian framework, the posterior distribution $p(f|\mathcal D,\theta)$ is used to make predictions. The hyperparameters, denoted by $\theta$, are determined by the maximizing the marginal distribution $p(\mathcal D|\theta)$. The benefit of Bayesian learning of $\theta$ is that over fitting is naturally avoided by the two competing terms, data-fit and complexity penalty, in the marginal likelihood.


Deep GP serves as a prior distribution over both {\it homogeneous} and {\it heterogeneous} compositions of functions, where the resultant function $f$ from $L$ layers of warping is given by
\begin{equation}
    {\bf f}\sim\mathcal{GP}(\mu^{L}, k^{(L)}({\bf h}^{L-1},{\bf h}^{L-1}))
\end{equation} with the hidden functions ${\bf h}^{1:L}$ 
\begin{equation}
    {\bf h}^{m}\sim\mathcal{GP}(\mu^{m},k^{(m)}({\bf h}^{m-1},{\bf h}^{m-1}))\:.
\end{equation} The zeroth layer variable ${\bf h}^0$ represents the input ${\bf x}\in R^D$. The prior mean functions $\mu^{1:L}$ and covariance function $k^{(1:L)}$ in the latent layers are predetermined. Homogeneous deep GPs compose functions with the same kernel, $k^{(i)}=k^{(j)}, \hspace{1mm} \forall i, j$, whereas heterogeneous deep GPs compose functions with different kernels $\exists i, j \text{ s.t. } k^{(i)} \neq k^{(j)}$. For a finite set of $\{f_i\}_{i=1}^N$ associated with the input $\{{\bf x}_i\}_{i=1}^N$, the joint distribution can be expressed as
\begin{equation}
    p({\bf f}, {\bf h}^{1:L}|{\bf X})=
    p({\bf f}|{\bf h}^{L})p({\bf h}^{L}|{\bf h}^{L-1})\ldots
    p({\bf h}^{1}|{\bf X})\:,\label{joint_DGP}
\end{equation} in which each distribution in the product on right hand side is Gaussian. The DGP is defined as the marginal distribution,
\begin{equation}
    p({\bf f}|{\bf X})=\int p({\bf f},{\bf h}^{1:L}|{\bf X})d{\bf h}^{1:L}\label{DGP}
\end{equation} over the composition function. However, the above marginal distribution is not Gaussian as the latent variables ${\bf h}^{1:L}$ appear in the inverse of covariance matrix. 

Although the marginalization over the latent variable $\bf h$'s in Eq.~(\ref{DGP}) is intractable, we have the following observations regarding the joint and posterior distributions of DGP, which are consistent with the sampling-based DGP in~\citep{havasi2018inference}.

\begin{lemma}
The joint distribution in Eq.~(\ref{joint_DGP}) is invariant under the transformation ${\bf h}^{i}\rightarrow-{\bf h}^{i}$, $1\leq i\leq L$, if the prior mean $\mu^{i}=0$. 
\end{lemma}

The proof proceeds by noting that the latent variable ${\bf h}^i$ appears only in the two distributions, $p({\bf h}^i|{\bf h}^{i-1})$ and $p({\bf h}^{i+1}|{\bf h}^i)$ in Eq.~(\ref{joint_DGP}). In the former distribution, the quadratic ${\bf h}^i$ in the exponent of Gaussian distribution permits the sign changing. In the latter distribution, the covariance function $k^{(i+1)}$ is also invariant because the distance $|h_a^i-h_b^i|$, $1\leq a,b\leq N$, and the distance to origin, $|h^i_a|$ is intact by the sign change. Therefore the joint distribution is left invariant. 

\begin{remark}
If a set of function values associated with each latent layer, $\{{\overline{\bf h}}^1,{\overline{\bf h}}^2,\cdots,{\overline{\bf h}}^L\}$, is found to correspond to the maximum of the posterior distribution $p({\bf h}^{1:L}|\mathcal D)\propto\int p({\bf f},{\bf h}^{1:L})p(\mathcal D|{\bf f})d{\bf f}$, with zero prior mean in each latent layer, then the rest of $2^L$ sets, $\{\pm{\overline{\bf h}}^1,\cdots,\pm{\overline{\bf h}}^L\}$, are also maximum of posterior distribution. Therefore, such DGP posterior distribution is multi-modal.
\end{remark} 

In the following, we shall consider the family of few-layer DGP. Since DGP allows heterogeneous composition, it is convenient to label DGP by the covariance functions. For example, SE[SC] stands for the DGP that $L=2$ and the covariance functions $k^{(2)}$ and $k^{(1)}$, respectively, are squared exponential and squared cosine functions in the output and input layers. The key finding in this paper is, despite the fact that marginalization in DGP inference is intractable, that the calculation of moments is tractable and analytic for some families of DGP. 

To illustrate, we first consider the two-layer DGP, $L=2$, and the prior mean at output layer is zero, $\mu^{(2)}=0$. The first moment of the marginal distribution in Eq.~(\ref{DGP}) is obtained as $\EX[f_i]=\int f_ip({\bf f,h}|{\rm X})d{\bf f}d{\bf h}$, and it follows from the zero-mean assumption that this first moment is zero. Likewise, the third moment as well as other odd order moments is zero. Moreover, the second moment is $\EX[f_if_j]$, which can be shown to be the expected value of output layer covariance function with respect to the mutivariate normal distribution associated with the input layer GP,
\begin{equation}
    \EX[f_if_j] = 
    \int d{\bf h}\  k^{(2)}(h_i,h_j)\mathcal N({\bf h}|\mu^{(1)},k^{(1)})\:.\label{2ndMoment}
\end{equation} As will be shown later, the second moment is useful when the DGP is approximated by GP where the effective covariance function is just the second moment.  Additionally, the fourth moment is
\begin{equation}
    \EX[f_if_jf_mf_l]=\sum\int
    d{\bf h}\  k^{(2)}_{ij}k^{(2)}_{ml}\mathcal N({\bf h}|\mu^{(1)},k^{(1)})\:\label{4thMoment}
\end{equation} where, according to the Isserlis' theorem~\citep{isserlis1918formula}, the sum is over all distinct ways of partitioning $i,j,m,l$ into pairs $ij$ and $ml$. $k^{(2)}_{ij}=k^{(2)}(h_i,h_j)$ is abbreviated for convenience. Although the marginal distribution for DGP in Eq.~(\ref{DGP}) is intractable, the fourth moments are able to reveal additional information about the true distribution and shed light on the validity of approximation scheme in this paper.

\section{Second Moment of DGP}\label{sec:second_moment}

We first consider second moments of the DGP distributions, SE[$\cdot$] and SC[$\cdot$], where the covariance function $k^{(1)}$ in the input layer is arbitrary but $k^{(2)}$ in the output layer is in the exponential family with zero prior mean. The analytic calculation of second moment is possible due to the marginal property of Gaussian which leads to the bivariate distribution, 
\begin{equation}
    \mathcal N(h_i,h_j|{\bf v},{\mathbb K})=\int d{\bf h}_{\backslash ij}\ \mathcal N({\bf h}|\mu,k),\label{bivariate}
\end{equation} where $d{\bf h}_{\backslash ij}$ denotes all but the integration with respect to $h_i$ and $h_j$. $\bf v$ and $\mathbb K$ denotes the mean and covariance matrix in the remaining block related to $h_i$ and $h_j$. Explicitly,
\begin{equation}
    {\mathbb K}=\left(
    \begin{array}{cc}
         k_{ii} &  k_{ij} \\
         k_{ij} &  k_{jj} 
    \end{array}
    \right)\:,
\end{equation} and the matrix elements take the value $k_{ij}=k({\bf x}_i,{\bf x}_j)$. As such, following the expression in Eq.~(\ref{2ndMoment}), the second moments of SE[$\cdot$] is obtained as
\begin{equation}
    \sigma^2\EX_{h_ih_j}\left[e^{-
    \frac{(h_i-h_j)^2}{2\ell^2}
    }\right]\:,\label{2nd_SE}
\end{equation} with the pair of random variables $h$'s sampled from the bivariate normal distribution in Eq.~(\ref{bivariate}). Similarly, the second moment of SC[$\cdot$] is given by
\begin{equation}
 \frac{\sigma^2}{2}\EX_{h_ih_j}\left(1+e^{
 i\frac{h_i-h_j}{\ell}
 }\right)\:.\label{2nd_SC} 
\end{equation} The signal magnitude $\sigma_2$ and length scale $\ell_2$ are the hyperparameters in the output layer GP.

\subsection{Squared Exponential DGP: SE[$\cdot$] \& NuN[$\cdot$]}

The following lemma is useful when calculating the expected value of random exponential quadratic function.

\begin{lemma}
The expected value of the random exponential quadratic form $Q({\bf x})=\exp(-\frac{{\bf x}^tJ{\bf x}}{2})$ with respect to multivariate normal distribution $\mathcal N({\bf x}|{\bf v},\mathbb K)$ is given by
\begin{equation}
    \EX[Q({\bf x})]=
    \frac{
    \exp(-\frac{1}{2}{\bf v}^t\mathbb A{\bf v})}
    {\sqrt{|I+{\mathbb K} J|}}
    \:,\label{SE[Nonstationary]}
\end{equation} where $J$ denotes the symmetric matrix of dimension $|{\bf x}|$ and the matrix $\mathbb A = \mathbb K^{-1}[I-(I+{\mathbb K}J)^{-1}]$. 
\end{lemma}

For SE[$\cdot$] with prior mean being zero in input and output layers, the calculation Eq.~(\ref{2nd_SE}) for covariance function in above representation with $J=\frac{1}{\ell^2}\bigl(\begin{smallmatrix}
1&-1\\ -1&1
\end{smallmatrix} \bigr)$ leads to the second moment,
\begin{equation}
    \sigma^2\left(
    1+\frac{k_{ii}+k_{jj}-2k_{ij}}{\ell^2}
    \right)^{-\frac{1}{2}}\:.\label{SESEKernel}
\end{equation} Moreover, we may consider the covariance function, $k(x,y)=\sigma^2\exp[-(\alpha x^2-2\beta xy+\alpha y^2)/2]$ with $\alpha>\beta>0$, as a more general Gaussian kernel in the output layer. This kernel in fact corresponds to the neural network with Gaussian activation function [Eq.~(13) in~\citep{williams1997computing}]. For convenience, we label such DGP as NuN[$\cdot$]. It should be noted that NuN[$\cdot$] is identical to SE[$\cdot$] as $\alpha=\beta=1/\ell^2$. Writing the matrix $J=\bigl(\begin{smallmatrix}
\alpha&-\beta\\ -\beta&\alpha
\end{smallmatrix} \bigr)$, the resultant second moment reads, 
\begin{equation}
    \sigma^2\left[
    1+
    \alpha(k_{ii}+k_{jj})-2\beta k_{ij}
    +(\alpha^2-\beta^2)|\mathbb K|
    \right]^{-\frac{1}{2}}\:.\label{Gaussian}
\end{equation} Note that the determinant term $|\mathbb K|=k_{ii}k_{jj}-k_{ij}^2$ does not appear in the stationary SE[$\cdot$] case. In addition, the nonstationariness, i.e. the covariance function value depending on the input ${\bf x}_i$ and ${\bf x}_j$ as the distance between pair vanishes, is intact as long as $\alpha\neq\beta$. Otherwise, Eq.~(\ref{Gaussian}) shall be constant as $k_{ij}\rightarrow k_{ii}$ and $k_{jj}$.


\subsection{Squared Cosine DGP: SC[$\cdot$]}
To capture periodic patterns in data, we may be interested in squared cosine (SC) kernel, $k(x,y)=\sigma^2\cos^2\frac{x-y}{2\ell}$. The expectation of general exponential linear function is given by the following lemma.
\begin{lemma}
The expected value of the random exponential linear function $L({\bf x})=\exp({\bf u}\cdot{\bf x})$ with respect to the bivariate nomral distribution $\mathcal N({\bf x}|{\bf v},\mathbb K)$ is given by
\begin{equation}
    \EX[L({\bf x})]=
    \exp(
    {\bf u}\cdot{\bf v}+
    \frac{
    {\bf u}^t\mathbb K{\bf u}
    }
    {2}
    )\:.
\end{equation}
\end{lemma} Using ${\bf u}=\frac{1}{\ell}\bigl(\begin{smallmatrix}
i\\ -i
\end{smallmatrix} \bigr)$, the second moment of SC[$\cdot$] with zero prior mean in input layer, Eq.~(\ref{2nd_SC}) becomes
\begin{equation}
	\frac{\sigma^2}{2}\left[1+\exp{\frac{2k_{ij}-k_{ii}-k_{jj}}{2\ell^2}}
    \right]\:.\label{CosKernel}
\end{equation}

\subsection{Three-layer DGP: SE[SE[SE]]}

For DGP with $L>2$, the present approach for obtaining the second moment is still valid but exact closed form seems unlikely. To be able to shed light on the effect of depth on the second moment, we consider the special case for three-layer DGP, SE[SE[SE]], where SE kernel is employed in all three layers and all prior means are set to zero. Following from the second moment of SE[SE], Eq.~(\ref{SESEKernel}), we can show that the second moment for three-layer SE[SE[$\cdot$]] is,
\begin{equation}
	\EX_{h_ih_j}\left\{
	\sigma_3^2[1+2\frac{\sigma_2^2}{\ell_3^2}(1-c)]^{-\frac{1}{2}}
	\right\}
	\label{2ndMoment_3layer}
\end{equation} with $c=\exp[-(h_i-h_j)^2/2\ell_2^2]$ and the $h$'s are sampled from the bivariate normal distribution $\mathcal N(0,k)$. We may proceed by approximating the random second moment inside the braces in Eq.~(\ref{2ndMoment_3layer}) as
\[
    \begin{dcases}
        \sigma_3^2 & |h_i-h_j|\leq \ell_2 \\
        \frac{\sigma_3^2}{\sqrt{1+2\frac{\sigma_2^2}{\ell_3^2}}} & {\rm otherwise}\:. \\
    \end{dcases}
\] As such, we obtain the following approximate second moment for three-layer DGP,
\begin{equation}
    \frac{\sigma_3^2}
    {\sqrt{1+2\frac{\sigma_2^2}{\ell_3^2}}}[1-{\rm erf}(v)]+
    \sigma_3^2\ {\rm erf}(v)\label{SESESE_1}
\end{equation} where erf() represents the error function and the parameter $v=\frac{\ell_2}{\sqrt{2(k_{ii}-k_{ij})}}$.

\section{Fourth Moment of SE[$\cdot$] and SC[$\cdot$]}\label{sec:fourth_moment}

The fourth moment is useful in revealing some important properties, e.g. heavy-tailedness, of a distribution. For univariate distribution, the measure of excess kurtosis is linked to the Gaussianity. For example, the heavy-tailed Student t-distribution has positive excess kurtosis whereas the normal distribution has zero. Here, we quantify the non-Gaussianity of DGP in Eq.~(\ref{DGP}) by investigating the fourth moments of the special cases SE[$\cdot$] and SC[$\cdot$]. 

\begin{lemma}
The product of two exponential quadratic functions has the expected value with respect to the multivariate distribution $\mathcal N(h_i,h_j,h_m,h_l|0,K)$, 
\begin{equation}
    \EX[e^{-(h_i-h_j)^2-(h_m-h_l)^2}]=[G_{ij}G_{ml}-V]^{-\frac{1}{2}}\label{4thMoment_SESE}
\end{equation} where $\frac{1}{\sqrt{G_{ij}}}=\EX[e^{-(h_i-h_j)^2}]$ corresponds to the second moment for SE[$\cdot$]. The term $V=(k_{im}+k_{jl}-k_{il}-k_{jm})^2$, with the $k$'s denoting the matrix elements of off-diagonal block of covariance matrix $K$. The positive $V$ leads to the inequality,
\begin{equation}
	\EX[e^{-(h_i-h_j)^2-(h_m-h_l)^2}]\geq\EX[e^{-(h_i-h_j)^2}]\EX[e^{-(h_m-h_l)^2}]\label{inequal_SE}
\end{equation}
\end{lemma}
The equality follows the generalization of Lemma 2 to the case of four random variables. The determinant $|I_4+KJ|$ is a bit tedious but can be done with the help of the equality, $|\bigl(\begin{smallmatrix}
A&C\\ D&B \end{smallmatrix} \bigr)|=|A||B||I_2-A^{-1}CB^{-1}D|$.    

\begin{lemma}
The expectation value of the following complex exponential function with respect to the above multivariate distribution is,
\begin{equation}
	\EX[e^{i(h_i-h_j)+i(h_m-h_l)}] = e^{2k_{ij}-k_{ii}-k_{jj}}e^{2k_{lm}-k_{ll}-k_{mm}}e^{2(k_{il}+k_{jm}-k_{im}-k_{jl})}\:,\label{4thMoment_SCSC}
\end{equation} and hence the inequality holds,
\begin{equation}
	\EX[\cos(h_i-h_j)\cos(h_m-h_l)]\geq\EX[\cos(h_i-h_j)]\EX[\cos(h_m-h_l)]\:.\label{inequal_SC}
\end{equation}
\end{lemma}
The equality is straightforward by applying Lemma 3. The detailed proof of the inequality can be found in Supplemental Material where we use the fact that the arithmetic mean is larger than or equal to the geometrical mean.

\section{Approximating DGP with GP}\label{sec:approximation}

In previous sections, we have shown the second and fourth moments of some DGP models. Despite the fact that DGP distribution in Eq.~(\ref{DGP}) is not tractable and non-Gaussian, these moments are exact and analytic (except the case of SE[SE[SE]]). Here, we propose to approximate the true DPG distribution with GP. Namely, the variational distribution,
\begin{equation}
    q = \mathcal N({\bf f}|0, k_{\rm eff})\label{approximate_q}
\end{equation} is a multivariate normal distribution which minimizes the KL divergence, ${\rm KL}(p||q)$ (this usage is not standard as the conventional KL is between true and approximate posterior distributions). Due to the Gaussianity of $q$, the second moments for true distribution $p$ in Eq.~(\ref{DGP}) have the exact correspondence,
\begin{equation}
    K_{\rm eff}=\EX_q[{\bf f}{\bf f}^t]=\EX_p[{\bf f}{\bf f}^t]\:.
\end{equation} Therefore, the second moments turn out to be the effective covariance function $k_{\rm eff}: R^D\times R^D\rightarrow R$. 

Considering the case of SE[SE], the approximate distribution $q$ acquires the effective covariance function,
\begin{equation}
    k_{\rm eff}({\bf x},{\bf y})=
    \frac{\sigma_2^2}
    {
    \sqrt{
    1+2\frac{\sigma_1^2}{\ell_2^2}
    [
    1-\exp(-||{\bf x}-{\bf y}||^2/2\ell_1^2)
    ]
    }
    }\:,\label{kernel_2layerSESE}
\end{equation} where the $\sigma$'s are signal magnitude and $\ell$'s the length scale in the respective layers. In addition, the heterogeneous composition SE[Lin] with linear kernel $k({\bf x},{\bf y})={\bf x}\cdot{\bf y}$ in input layer gives rise to the rational quadratic kernel of order $\frac{1}{2}$. The composition can also include the non-stationary kernel such as the neural network kernel~\citep{williams1997computing} in the first layer. 

Likewise, the composition SC[$\cdot$] results in many novel and interesting kernels as well. Composition SC[SE] generates an exponential Gaussian kernel. Surprisingly, the periodic kernel~\citep{mackay1998introduction} is identical to the composition SC[SC], and SC[Lin] turns out to be the addition of SE and constant kernels. See Supplementary Material for effective covariance function of SC[NuN].

For the two-layer composition SE[$\cdot$], it is clear that the first layer can be regarded as a deterministic mapping when the signal magnitude $\sigma_1$ is set to zero. Thus, with $\mu^{(1)}\neq0$, this two-layer DGP is equivalent to $\mathcal{GP}(0,k(\mu^{(1)}(\cdot),\mu^{(1)}(\cdot)))$. The nature of DGP distribution becomes transparent by inspecting the fourth moment. The fourth moment of the approximate distribution for SE[$\cdot$], 
\begin{equation}
    \EX_q[f_i^2f_j^2]=\sigma_2^4
    \{
    1+[1+(k_{ii}+k_{jj}-2k_{ij})/\ell_2^2]^{-1}
    \}\:,\label{ex_4thmoment_q}
\end{equation} whereas the true distribution,
\begin{equation}
    \EX_p[f_i^2f_j^2]=\sigma_2^4\{1+[1+2(k_{ii}+k_{jj}-2k_{ij})/\ell_2^2]^{-1/2}\}\:.\label{ex_4thmoment_p}
\end{equation} If we treat the terms inside the bracket as $1+x$ with $x>0$, following the fact that $1/\sqrt{1+2x}\geq1/(1+x)$ for $x\geq0$, then Eq.~(\ref{ex_4thmoment_p}) is larger than or equal to Eq.~(\ref{ex_4thmoment_q}). Nevertheless, for large value of $\ell_2^2$, Eq.~(\ref{ex_4thmoment_q}) and (\ref{ex_4thmoment_p}) are identical up to $O(\ell_2^{-2})$ in the expansion. Therefore, the distributions $q$ for SE[SE] is a good approximation to $p$ in the regime $\sigma_1/\ell_2\ll1$. More generally, the following remark states the fourth moments from $p$ and $q$ and the nature of distribution of SE[$\cdot$].

\begin{remark}
The fourth moments of SE[$\cdot$] and SC[$\cdot$] associated with the true distribution $p({\bf f}|{\bf X})$ in Eq.~(\ref{DGP}) and the approximate distribution $q({\bf f}|{\bf X})$ in Eq.~(\ref{approximate_q}) have the relation,
\begin{equation}
    \EX_p[f_if_jf_mf_l]\geq\EX_q[f_if_jf_mf_l]\:,
\end{equation} provided that their second moments match $\EX_p[f_if_j]=\EX_q[f_if_j]$.
\end{remark} 
The proof follows from that the mutivariate normal distribution has $\EX_q[f_if_jf_mf_l]=\sum\EX_q[f_af_b]\EX_q[f_cf_d]$ with the summation over all three distinct ways of partitioning the quartet $\{i,j,m,l\}$ into two doublets~\citep{isserlis1918formula}. As for $p$, the partition is the same but each term is instead given by Eq.~(\ref{4thMoment_SESE}) (SE[ ]) and Eq.~\ref{4thMoment_SCSC} (SC[ ]), which are both greater than the product or the second moments (See the inequalities in Eq.~(\ref{inequal_SE}) and Eq.~(\ref{inequal_SC})). The particular fourth moment considered in Eq.~(\ref{ex_4thmoment_q}) and (\ref{ex_4thmoment_p}) is reproduced by setting the random variables $f_i=f_m$ and $f_j=f_l$ here.


An important implication of above remark is that the true distribution for this particular DGP might be a heavy-tailed non-Gaussian distribution. The above analysis is analogous to the comparison between GP and Student-t process (TP)~\citep{shah2014student}. A TP with same covariance function as a GP can be shown to possess larger fourth moment than the GP~\citep{press2005applied}, which may account for the more extreme behavior in TP.   


\subsection{Recursive Composition}

The above composition for two-layer DGP can be generalized to the cases where $L>2$ in a recursive manner. For \textit{homogeneous} composition, e.g. SE[SE[SE]]], one can treat SE[SE] inside the bracket as a GP with zero mean and effective covariance function given in Eq.~(\ref{kernel_2layerSESE}). Thus, the new effective covariance function for the three-layer DGP is obtained by  
plugging this kernel back into Eq.~(\ref{SESEKernel}). This trick can be applied recursively so in general the following relation holds,
\begin{equation}
    k^{(L+1)}_{\rm eff}({\bf x}_i, {\bf x}_j)=\frac{\sigma_{L+1}^2}{\sqrt{1+2(\ell_{L+1}^{-2})[\sigma_L^2-k^{(L)}_{\rm eff}({\bf x}_i, {\bf x}_j)]}}\:,
\label{eq:recursion}
\end{equation} which relates the current covariance function $k^{(L)}$ to the covariance $k^{(L+1)}$ after another layer of composition. Similar approach can be taken for obtaining other homogeneous composition kernels SC[SC[SC[...]]] (Eq.~\ref{CosKernel}) and the non-stationary Gaussian kernels (Eq.~\ref{Gaussian}). 

For \textit{heterogeneous} composition, e.g. SE[SC[NuN]], one can obtain the effective covariance function in similar manner, which leads to the expression for the effective covariance function,
\begin{equation}
    \frac{\sigma_3^2}{
    \sqrt{1+
    2\frac{\sigma_2^2}{\ell_3^2}
    [1-e^{(2k_{ij}-k_{ii}-k_{jj})/2\ell_2^2}]}
    },
\end{equation} where the NuN covariance function $k$ can be the general Gaussian kernel or the arcsine kernel [Eq.~(11) in~\citep{williams1997computing}].

\subsection{Characteristics of Effective Kernel} 

Here we consider the distribution over derivatives of composite functions sampled from approximate DGP in order to make connection with the expressivity parameter defined in NN model~\citep{Poole2016ExponentialEI}.
By definition, any two values $f_1$ and $f_2$ from a function $f(x)$ must follow the joint Gaussian distribution $\mathcal {N}(f_1,f_2|0,\Sigma)$ with the two-by-two covariance matrix $\Sigma$. In order to shed light on the expressiveness of our effective kernels, we calculate the following expectation,
\begin{equation}
    \EX[(f_1-f_2)^2] = \int df_1 df_2 (f_1-f_2)^2 \mathcal{N}(f_1,f_2|0,\Sigma)\:,
\end{equation} which shall asymptotically approach $\mathscr{E}\{[f'(x)]^2\}(x_1-x_2)^2$ as their the inputs $x_1$ and $x_2$ are close to each other. Consequently, we obtain the expected value of squared derivative, 
\begin{equation}
    \EX\{[f'(x)]^2\}= 2\lim_{x_1\rightarrow x_2} \frac{\sigma_2^2 - k_{\rm eff}(x_1, x_2)}{(x_1-x_2)^2}\:.
\end{equation} When applying to the effective kernel of SE[SE] in Eq.~(\ref{SESEKernel}), one may show the normalized and squared average derivative has 
\begin{equation}
    \frac{\EX\{[f'(x)]^2\}}{\sigma_2^2} = \frac{\sigma_1^2}{\ell_2^2\ell_1^2} = \frac{\chi}{\ell_1^2}\:,
\end{equation} where we follow~\cite{Poole2016ExponentialEI} and define the parameter important for the analysis of expressivity,
\begin{equation}
    \chi:=\frac{\partial (k_{\rm eff}/\sigma_2^2)}{\partial (k/\sigma_1^2)}\rvert_{k=\sigma_1^2}\:,\label{chi}
\end{equation} which was suggested to signal the phase transition between chaotic and ordered phases in neural network~\citep{sompolinsky1988chaos,Poole2016ExponentialEI}. With the chain rule for derivative, we can also show that the effective SE[SE[SE]] has the normalized and squared average derivative $\chi/\ell_1^2$ with $\chi=\sigma_2^2\sigma_1^2/\ell_3^2\ell_2^2$. Similar construction is applicable to squared cosine DGP. This characteristic of the derivative of our effective kernels is consistent with~\citep{duvenaud2014avoiding}.



\section{Interpretation of Effectively DGP}\label{sec:interpretation}


Our approximation based on GPs yields benefits in interpretability for both DGPs and DNNs. 
Prior results establish that DNNs can be viewed as GPs~\citep{williams1997computing,cho2009kernel,lee2017deep}. 
In both the DGP and DNNs paradigms, people tend toward homogeneous kernels / activation functions, in part because there are not effective tools for predicting how compositions will behave a priori. 
In Table~\ref{kerneltable}, we list a few representative analytic kernels from stacking two GPs. Our analysis allows iterative generation of heterogeneous kernels such as SE[SC[NuN]] and homogeneous one SE[SE[SE[...]]] of any depth. 
Moreover, because our approach sits between these two paradigms, we can view our compositions as arising from activation functions~\citep{Daniely2016TowardDU}, kernels~\citep{scholkopf1998nonlinear,duvenaud2013structure}, or compositions of both.

\begin{table}[t] 
\begin{center}
\begin{tabular}{ll}
\multicolumn{1}{l}{\bf Notation}  &\multicolumn{1}{l}{\bf Effective kernel} \\
\hline 
SE[SE]         &$a[1+b\ G(\Delta x^2,c)]^{-\frac{1}{2}}$ \\
SC[SE]             &$a\{1+\exp[-bG(\Delta x^2,c)]\}$ \\
SE[SC]             &$a[1+b\sin^2(\Delta x/c)]^{-\frac{1}{2}}$\\ 
SC[SC]   &$a\{1+\exp[-b\sin^2(\Delta x/c)]\}$\\
SE[Lin] & RQ${\frac{1}{2}}$\\
SC[Lin] & SE\ +\ Const.\\
SE[Lin+SE] & $a\{1+bG(\Delta x^2,c)+d\Delta x^2\}^{-\frac{1}{2}}$ \\
NuN[SE] & $a[1+f+bG(\Delta x^2, c/2)+dG(\Delta x^2, c)]^{-\frac{1}{2}}$
\end{tabular}
\vspace{0.1cm}
\caption{\label{kerneltable} List of compositional kernels by stacking two GPs with respective kernels denoted by SE (squared exponential), SC (squared cosine), Lin (linear), and NN (neural network with Gaussian activation function). The function symbol $G(x^2,c)=1-\exp(-x^2/c)$ appears with SE composition. The sequence can be understood from the notation, for example, SE[SC] means input data is directed to the first GP with SC kernels, followed by sending its output to the second GP with SE kernel which produces the final output. The hyperparameters are represented by the symbols $a$, $b$, $c$, $d$, and $f$, all of which are positive.}
\end{center}
\end{table}




We may leverage interpretability to analyze differences in expressivity of SE[SE] and SE. First, SE[SE] approaches RQ$\frac{1}{2}$ in small $\Delta x$ limit. Because the rational quadratic kernel is known to arise from summing over infinitely many SE kernels with Gamma distribution over the inverse of length scale~\citep{rasmussen2006gaussian}, the effective SE[SE] kernel also possesses the multi-length scale feature. Second, the function composition $y=f(h(x))$ leads to the long-range correlation, i.e. when $\Delta x\rightarrow\infty$, SE[SE] kernel approaches some nonzero constant while SE becomes vanishingly small. This can be seen by noting that the first-layer outputs $h(x_1)$ and $h(x_2)$ are nearly independent if $x_1$ and $x_2$ are far apart. Nevertheless, the distance $h(x_1)-h(x_2)$, generally falling within $[0, \sigma_1]$ with high probability, fed into the second layer is greatly reduced if the signal magnitude $\sigma_1$ associated with $h$ is small, which leads to finite correlation between $f(h(x_1))$ and $f(h(x_2))$.


The composition using non-stationary kernel, i.e. NuN[$\cdot$], is also of interest. 
Take NuN[SE] for example, the determinant $|\mathbb K|$ in Eq.~(\ref{Gaussian}) generate a squared term, $k_{ij}^2$, and as such the combination $e^{-||{\bf x}_i-{\bf x}_j||^2/\ell^2}+e^{-||{\bf x}_i -{\bf x}_j||^2/2\ell^2}$ appears in the effective covariance function.

\subsection{SE vs.~SE[SE]}

\begin{figure}[ht]
  \centering
  \includegraphics[width=0.8\columnwidth]{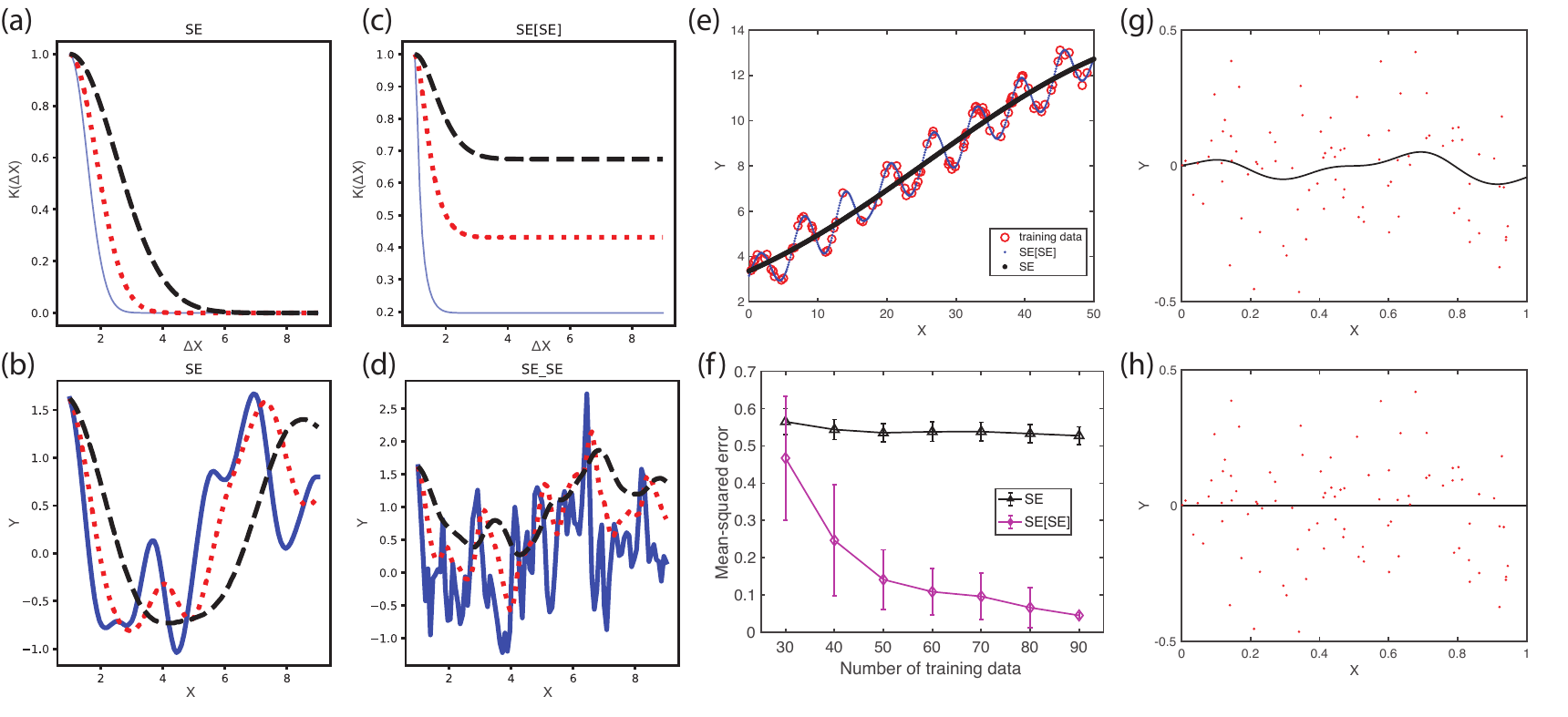}
  \caption{\textbf{(a)} Three different SE kernels and \textbf{(b)} functions sampled from these kernels. \textbf{(c)} Three different SE[SE] kernels and \textbf{(d)} functions sampled from these kernels. For (a) and (c), the kernel black dash, red dotted line, and blue solid line are 0.8, 0.5, and 0.2 at $\Delta x = 1$, respectively. All functions in (b) and (d) are generated using the same noise vector. \textbf{(e)-(f)} Regression on multiscale data with SE and SE[SE] kernels. \textbf{(g)-(h)} Regression with SE (g) and SE[SE] (h) kernels on pure noise data sampled from normal distribution with $\sigma=0.2$. Red dots are data points, and black line is the predictive mean.}
  \label{SE_vs_SESE}
\end{figure}

Figure~\ref{SE_vs_SESE}(a)--(d) shows functions generated from SE and SE[SE]. The functions generated by SE[SE] seem to possess two length scales in variations: the rapidly varying blue line in Figure~\ref{SE_vs_SESE}(d) seems to have slowly varying underlying trend that is similar to a shifted version of the black dash line. This is broadly consistent with the discussion in Section~\ref{sec:interpretation} of the additional expressivity of multi-length scale behavior in deeper structure.

To further demonstrate the expressivity of SE[SE] kernel over the SE kernel, we use them to fit a dataset with two length scales.\footnote{The data is from MATLAB FITRGP example. 
} Figure~\ref{SE_vs_SESE}(e) shows that the smooth SE kernel does not fit the small variation, while the SE[SE] kernel is able to capture these fast variation on top of the slowly varying one. Figure~\ref{SE_vs_SESE}(f) shows that the SE[SE] kernel is better than SE kernel at prediction not only in the range of larger amount of data, but also in the limited data regime. We also remark in passing that the prediction accuracy for SE[SE] kernel is comparable with the RQ kernel, which is known to be multi-length scale. 

Given the ability of the SE[SE] kernel to capture fast variation, one might have the concern that it will fit to noise. To explore this possibility, we trained both SE and SE[SE] kernels on pure noise data. In Figure~\ref{SE_vs_SESE}(g) and (h), we show the prediction mean (black) from training on 90 noise data points sampled from the normal distribution with $\sigma=0.2$. For this particular noise data, the SE kernel in (g) generates a non-zero prediction mean, while the SE[SE] kernel in (f) nicely predicts the underlying zero function.

We also apply the SE and SE[SE] kernels to two UCI regression data sets, the House Price dataset and the Abalone dataset. We investigated the test error as a function of the fraction of training number. For the House Price dataset, the SE[SE] kernel obtains lower test error than the SE kernel does when the fraction of training data is larger than 30\% (see Supplementary Material Figure 1(a)). For the Abalone dataset, the two kernels have similar test error (see Supplementary Material Figure 1(b)). 

\subsection{Deep SE Compositions}

\begin{figure}[t]
  \centering
  \includegraphics[width=0.8\columnwidth]{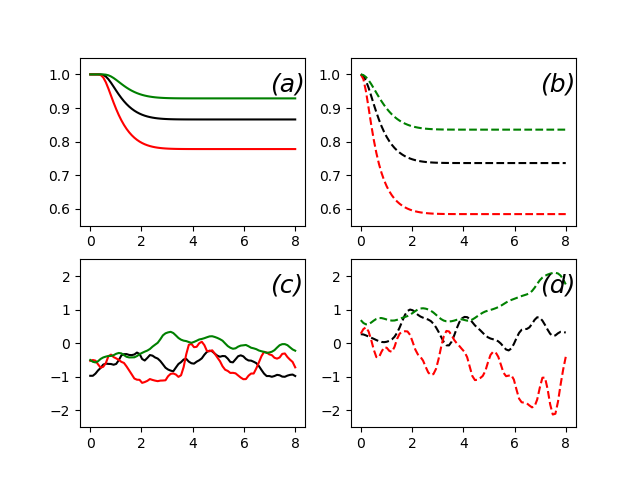}
  \caption{Sampling functions from effective kernels for SE[SE[SE]]. Panels (a) refers to the kernel using the approximate second moment in Eq.~(\ref{SESESE_1}) while panel (b) uses the recursive relation in Eq.~(\ref{eq:recursion}). The hyperparameters: green ($\sigma_2/\ell_3=0.8$, $\sigma_1/\ell_2=0.8$), black ($\sigma_2/\ell_3=1$, $\sigma_1/\ell_2=1$), and red ($\sigma_2/\ell_3=1.4$, $\sigma_1/\ell_2=1.2$). The common parameters: $\sigma_3=\ell_1=1$. Panel (c) and (d) shows the functions sampled from panel (a) and (b), respectively.}
  \label{SESESE}
\end{figure}

We examine two different methods for obtaining the effective covariance function in SE[SE[SE]]. The straightforward way deals with the expectation of second moment from output to input layers. With  approximation, Eq.~(\ref{SESESE_1}) is obtained. On the other hand, the recursive approach treats the first two layers SE[SE] inside the bracket as a GP first, then take the three-layer DGP as another two-layer DGP, which results in Eq.~(\ref{eq:recursion}). Fig.~\ref{SESESE} (a) and (b), respectively, demonstrates the effective kernel functions from Eq.~(\ref{SESESE_1}) and Eq.~(\ref{eq:recursion}) for different hyperparameters. The prominent difference is the presence of plateau for small $\Delta x$ in Fig.~\ref{SESESE}(a). For smaller value of $\sigma_2/\ell_3$ and $\sigma_1/\ell_2$, the two appraoches have closer results (green solid and dashed curves). The functions sampled from these kernels are shown in panel (c) and (d), respectively. With larger value of $\chi=(\frac{\sigma_2\sigma_1}{\ell_3\ell^2})^2$, the sampled functions in both panels seem to possess variation in shorter length scale. However, the sampled functions in panel (c) are more restricted in the range $[-\sigma_3,\sigma_3]$, but they are much less smooth than the functions obtained using kernel in (b).


\section{Hyperparameter Optimization and Expressivity}\label{sec:optimization}


From above analysis, it is clear that $\chi$ characterizes both the decay of kernel function [Fig.~\ref{SESESE} (a) and (b)] and the degree associated with variation in sampled functions. Considering the recursive kernel of depth $L$, $\chi\propto(\frac{\sigma}{\ell})^{L-1}$ increases exponentially with the $L$ if the ratio $\frac{\sigma_l}{\ell_{l+1}}$ is kept constant in the model.
In~\citep{Poole2016ExponentialEI}, the parameter defined as $\chi_1:=\partial c^l_{12}/\partial c^{l-1}_{12}$ characterizes the the mapping of manifold in DNNs and serves 
as expressivity parameter which also signals the transition between chaotic and ordered phases in a mean-field theory for neural network~\citep{sompolinsky1988chaos}.

\begin{figure}[t]
  \centering
  \includegraphics[width=0.8\columnwidth]{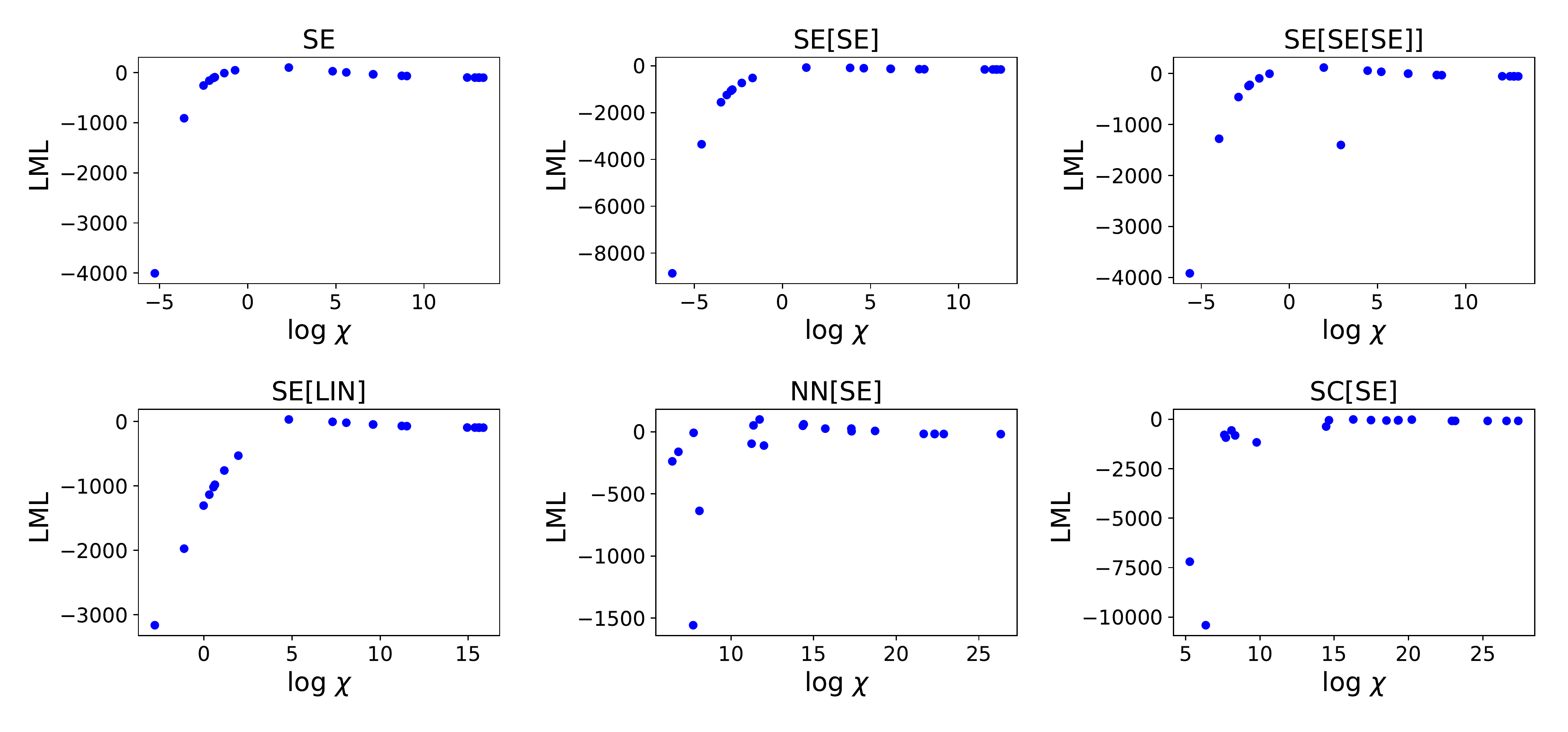}
  \caption{
  Log marginal likelihood (LML) for three-layer DGP SE[SE[SE]] [Eq.~(\ref{eq:recursion})] fitting the data generated by the kernel labeled on top of panel. The hyperparameters in generating kernels are chosen randomly. The LML shows a dramatic increase when the value of $\log\chi$, defined in Eq.~(\ref{chi}), is close to zero in the top row of data. This agrees with \cite{Poole2016ExponentialEI} who used $\chi$ from correlation maps to quantify the transition into the chaotic phase in which DNNs are much more expressive than shallow networks
  }
  \label{hyperparameter_optimization}
\end{figure}

As a demonstration that DGP with larger $\chi$ is more expressive, we generate data from the non-periodic kernels in Table~\ref{kerneltable}. The hyperparameters of the data-generating kernel are chosen randomly, and we use SE[SE[SE]] kernel to fit the data by maximum marginal likelihood. We run the optimization 20 times with 20 random initialization where the hyperparameters are sampled uniformly from (-10,10) in log space. Figure~\ref{hyperparameter_optimization} shows the final log of marginal likelihood versus the value of $\log\chi$ after optimization. It can be seen that a dramatic increase of marginal likelihood is observed near $\chi=O(1)$ in the top row of panels, while the transition is shifted in the bottom row. Therefore, the three-layer DGP have superior expressivity in the chaotic regime.

\section{CONCLUSIONS}\label{sec:conclusion}

We have presented interpretable deep Gaussian Processes that combine increased expressiveness associated with deep NNs with uncertainty quantification of GPs. Marginalization over latent function variable is not tractable but calculation of moments is. Our approach is based on approximating deep GPs as GPs, which enables one to analytically integrate yielding effectively deep, single layer kernels. The heavy-tailed nature of some deep Gaussian process distribution is revealed by the fourth moment. We have provided a recipe for constructing effective kernels for cases including homogeneous and heterogeneous kernels (equivalently, activation functions), derived a variety of such kernels, analyzed their behavior, and confirmed behavior by prior and posterior predictive simulation. Simpler than alternative approaches to variational inference, our approach yields strong benefits in interpretability while retaining remarkable expressivity.



\bibliographystyle{apalike}

\end{document}